\documentclass[oribibl]{llncs}
\pdfoutput=1
\usepackage{amsmath}
\usepackage{authblk}
\usepackage{subcaption}
\usepackage{geometry}
\usepackage{float}
\usepackage{amsfonts,amssymb}
\usepackage{subfiles}
\usepackage[colorlinks=true,allcolors=blue]{hyperref}
\usepackage{graphicx}
\graphicspath{{./figs/}}
\usepackage[
	style=numeric,
	maxnames=1,
    backend=bibtex8,
    hyperref=false,
    doi=false,
    isbn=false,
    url=false
]{biblatex}
\begin{document}
    
    \title{Generating Magnetic Resonance Spectroscopy Imaging Data of Brain Tumours from Linear, Non-Linear and Deep Learning Models}
    \author{Review Paper}
    \institute{Review Papers}
    \author{Nathan Olliverre\inst{1} \and Guang Yang\inst{3, 4} \and Gregory Slabaugh\inst{1} \and Constantino Carlos Reyes-Aldasoro\inst{2} \and Eduardo Alonso\inst{1}}
    \institute{Department of Computer Science, City University of London, EC1V 0HB, London, UK \and Department of Electrical Engineering, City University of London, EC1V 0HB, London, UK \and Cardiovascular Biomedical Research Unit, Royal Brompton Hospital, SW3 6NP, London, UK \and National Heart \& Lung Institute, Imperial College London, SW7 2AZ, London, UK}
    \maketitle
    \begin{abstract}
\label{sec:abstract}
Magnetic Resonance Spectroscopy (MRS) provides valuable information to help with the identification and understanding of brain tumors, yet MRS is not a widely available medical imaging modality. Aiming to counter this issue, this research draws on the advancements in machine learning techniques in other fields for the generation of artificial data. The generated methods were tested through the evaluation of their output against that of a real-world labelled MRS brain tumor data-set. Furthermore the resultant output from the generative techniques were each used to train separate traditional classifiers which were tested on a subset of the real MRS brain tumor dataset. The results suggest that there exist methods capable of producing accurate, ground truth based MRS voxels. These findings indicate that through generative techniques, large datasets can be made available for training deep, learning models for the use in brain tumor diagnosis.
\end{abstract}

    \par
    
    \section{Introduction} \label{sec:intro}
    
    Within the UK over 11,000 brain tumor cases are diagnosed each year~\cite{cancer_research_uk_brain_2015}. The survival rate and period of progression have been shown to improve with treatment~\cite{zacharaki_classification_2009}. The process to determine treatment can be difficult and time consuming~\cite{yang_discrete_2015}, this is further complicated due to limited numbers of staff available to perform these tasks~\cite{grant_overview:_2004}. These tasks also lead to the most errors in diagnosis~\cite{grant_overview:_2004}. The automation of the tasks involved in diagnosis could help to increase the accuracy and reduce the time it takes for the application of treatment to a patient.
    
\begin{figure}
        \centering
        \includegraphics[width=\linewidth]{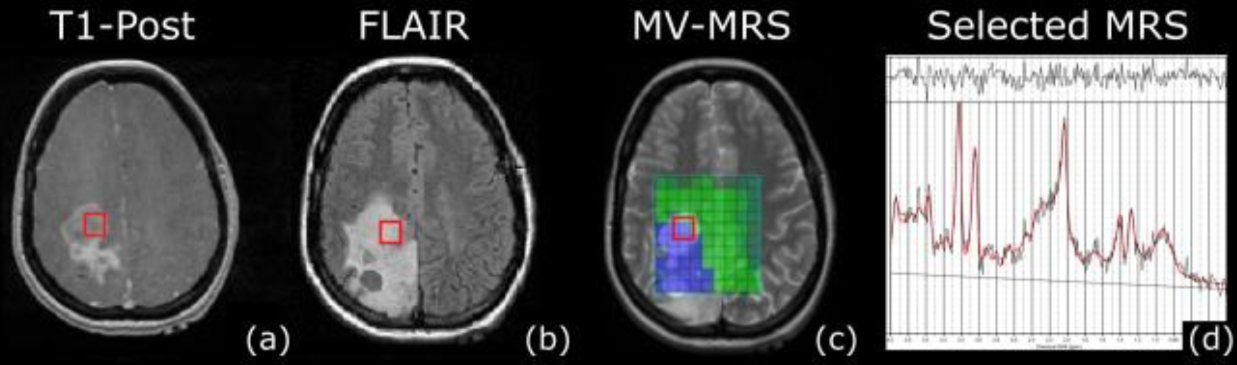}
        \caption{Example of a T1-Post MRI (a), Fluid-Attenuated Inversion Recovery (FLAIR) MRI (b), Multi-Voxel MRS Heatmap overlaid on FLAIR (c) and the selected Single-Voxel MRS highlighted in each of the images as the red square (d).}
        \label{fig:mri_mrs}
    \end{figure}

    Magnetic Resonance Spectroscopy (MRS), also known as MR Spectroscopic Imaging (MRSI) or Chemical Shift Imaging (CSI), provides a non-invasive method for the diagnosis of human tissue such as lung, bone or brain matter. Similar to MR Imaging (MRI), MRS is based upon the principles of Nuclear Magnetic Resonance (NMR)~\cite{andrew_nuclear_1992}; however, whereas MRI uses the resultant proton signals to create detailed graphical output, MRS uses the signals to determine the quantity in parts per million (ppm) of various metabolites within cells~\cite{gujar_magnetic_2005} which can be seen in the example shown in Figure~\ref{fig:mri_mrs}. One area in which MRS has been shown to provide valuable insights is in the case of brain tumors~\cite{howe_1h_2003}. In medical imaging research, MRS has proven to produce accurate results in identification of tumor grade classification~\cite{sibtain_clinical_2007,howe_1h_2003}.
    
    Current state-of-the-art computer vision models in medical imaging have begun to utilize the advancements made in machine learning, specifically in the field of deep learning~\cite{litjens_survey_2017}. These deep learning models take advantage of multi-layered networks to extract feature information from the input data. To power the ability for these deep models to extract such feature-rich information, large swaths of data are required~\cite{lecun_deep_2015}. This can be an obstacle in the medical imaging field as restrictions in data protection~\cite{cios_uniqueness_2002} along with non-standardized practices makes it difficult to collect the required amount of data to work with deep learning models. To be able to apply current state-of-the-art classifiers to MRS brain tumor images more data is required, hence techniques to fabricate or generate data are essential.
    
    Advancements in computer vision and machine learning have also given the rise to accurate generative techniques for the creation of artificial data. One such generative technique is Generative Adversarial Networks (GANs)~\cite{goodfellow_generative_2014} which have been developed to help create more domain specific/accurate artificial data. GANs use multiple models working against each other to create accurate data, with one model (Generator) attempting to produce artificial data capable of “fooling” a model trained to determine “real” from “fake” data (Discriminator). By using approaches such as GANs the process of creating data that adheres to the domain can be achieved. Recently, variations of the original GAN model have been developed to produce better results. The Deeply Convolutional GAN (DCGAN)~\cite{radford_unsupervised_2015}, which takes into account the improvements that deep learning models have shown against their traditional counterparts uses a multi-convolutional layered Generator and Discriminator. Although GANs have been shown to produce accurate results in various fields~\cite{radford_unsupervised_2015,salimans_improved_2016} they are known to be unstable and hard to train~\cite{arjovsky_towards_2017} whereas their more linear counterparts are considered to be easier to train but less expressive.
    
    This paper applies three of the state-of-the-art methods in generating synthetic data (GAN, DCGAN and a modified MRS brain tumor classifier~\cite{olliverre_pairwise_2017}) to the domain of MRS for review. To determine the accuracy of the artificial data created each generated dataset was used to train a Random Forest for the classification of brain tumors. The results from the trained Random Forest classifiers were bench-marked against the results of one trained on the real MRS images.

    
    \section{Materials \& Methods}\label{sec:methodsandmaterials}
    
    \subsection{Materials}
    
\begin{figure}
    	\centering
        \subcaptionbox{Healthy Data. \label{fig:dataset_healthy_data}}{\includegraphics[width=0.3\linewidth]{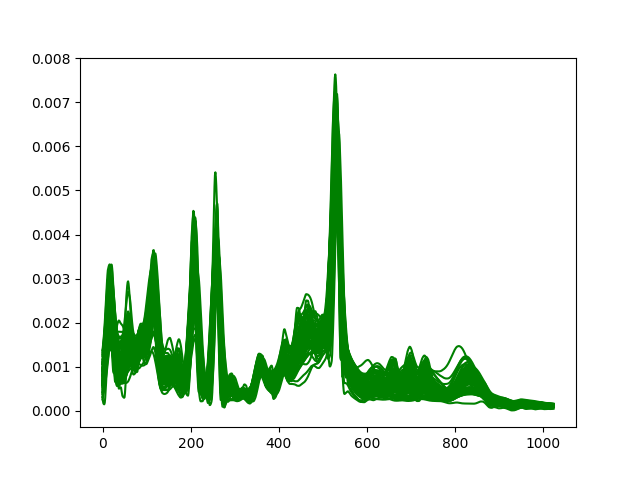}}
    	\subcaptionbox{Low Grade Data. \label{fig:dataset_low_data}}{\includegraphics[width=0.3\linewidth]{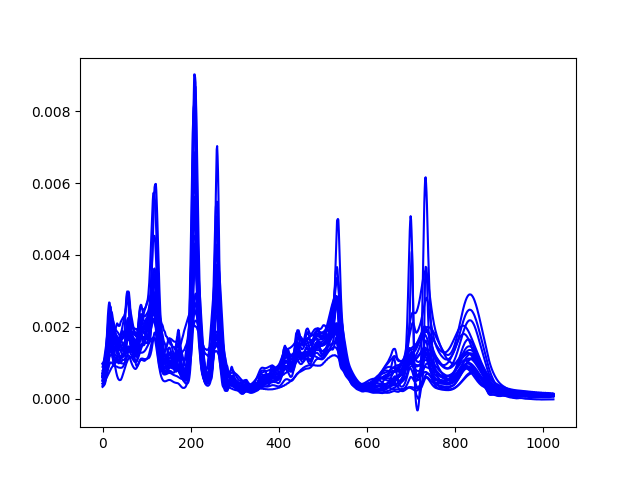}}
    	\subcaptionbox{High Grade Data. \label{fig:dataset_high_data}}{\includegraphics[width=0.3\linewidth]{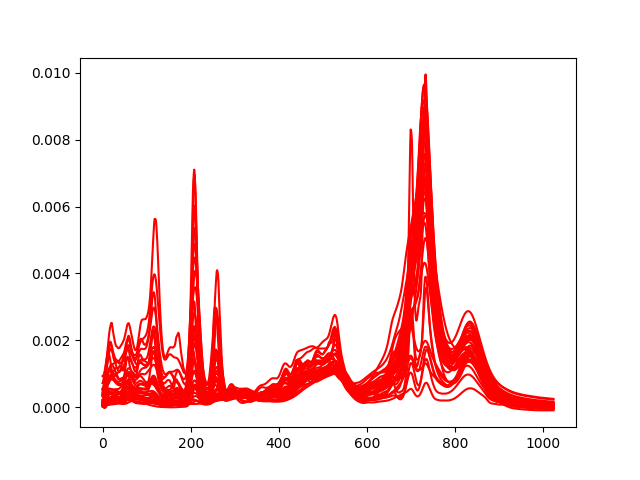}}
        \caption{Illustration of the
        MRS spectra grouped into \ref{fig:dataset_healthy_data} healthy (green), \ref{fig:dataset_low_data} low grade (blue) and \ref{fig:dataset_high_data} high grade (red) tumor tissue. Peaks correspond to metabolites in the region of interest.} \label{fig:dataset}
    \end{figure}

    The MRS dataset used in this study was obtained by St. George's University London and consisted of a single-voxel MRS training and testing set. Both the test and training sets were acquired using a GE Signa Horizon 1.5T MR system with a Repetition Time (TR) and a short Echo Time (TE) of 2000ms and 30ms respectively. A Point-Resolved spectroscopic sequence protocol was used to acquire the training and test dataset. The World Health Organization classifies tumours into 4 grades (WHO)~\cite{centre_international_de_recherche_sur_le_cancer_pathology_2004}: Grades I - IV, with GI and GII being deemed low grade tumors and GIII and GIV said to be high grade, malignant tumors. The composition of the training and test dataset was 137 samples. Of the training set 70 were classified as healthy tissue, 20 as GII (low grade), 10 as GIII and 20 GIV (high grade). The test set consisted of 9 healthy, 3 low grade and 5 high grade which were taken through random sampling of the entire dataset prior to training. Due to the similarity between GIII and GIV tumour tissue in MRS imaging, the GIII data was merged with the GIV data by labelling each as high grade. Figure~\ref{fig:dataset} illustrates MRS spectra of different grades. 
    
    The positioning for every scan captured in the dataset was placed on a homogeneous, representative tumor region determined by an expert using post-Gd contrast T1w, T2w and FLAIR structural contrast images alongside the relevant histopathological information. This was to ensure accuracy within the training data and that there was a heterogeneity of MRS characteristics represented within each voxel scanned. The individual labels for the data were achieved via the diagnosis of a biopsy by a practiced physician in which the clinical, radiological and histopathological information of each patient was incorporated to the diagnosis.
    
\begin{figure}
    	\centering
        \subcaptionbox{PCA applied with two components. \label{fig:datavis_pca}}{\includegraphics[width=0.45\linewidth]{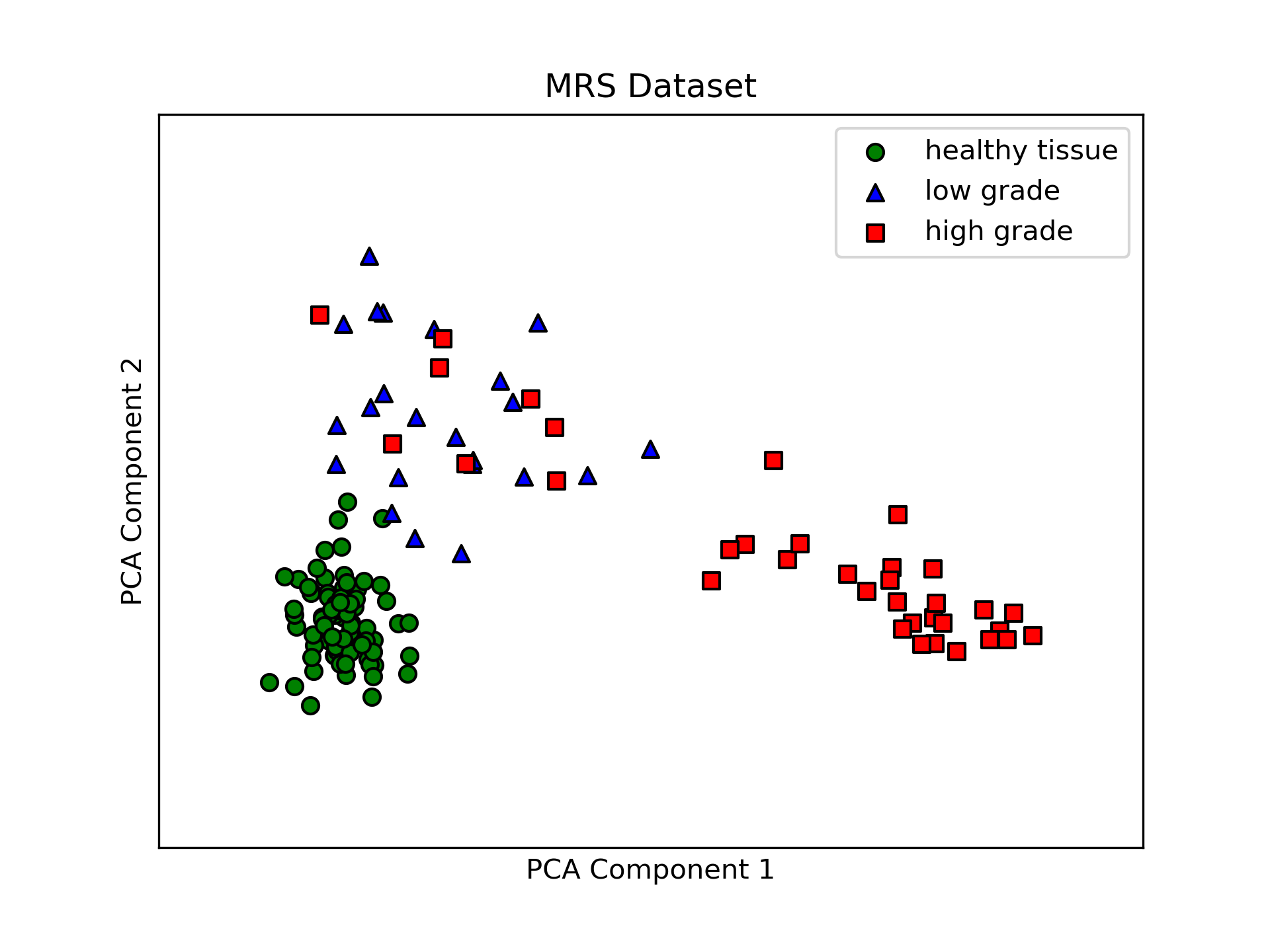}}
        \subcaptionbox{K-means clustering applied. \label{fig:datavis_kmeans}}{\includegraphics[width=0.45\linewidth]{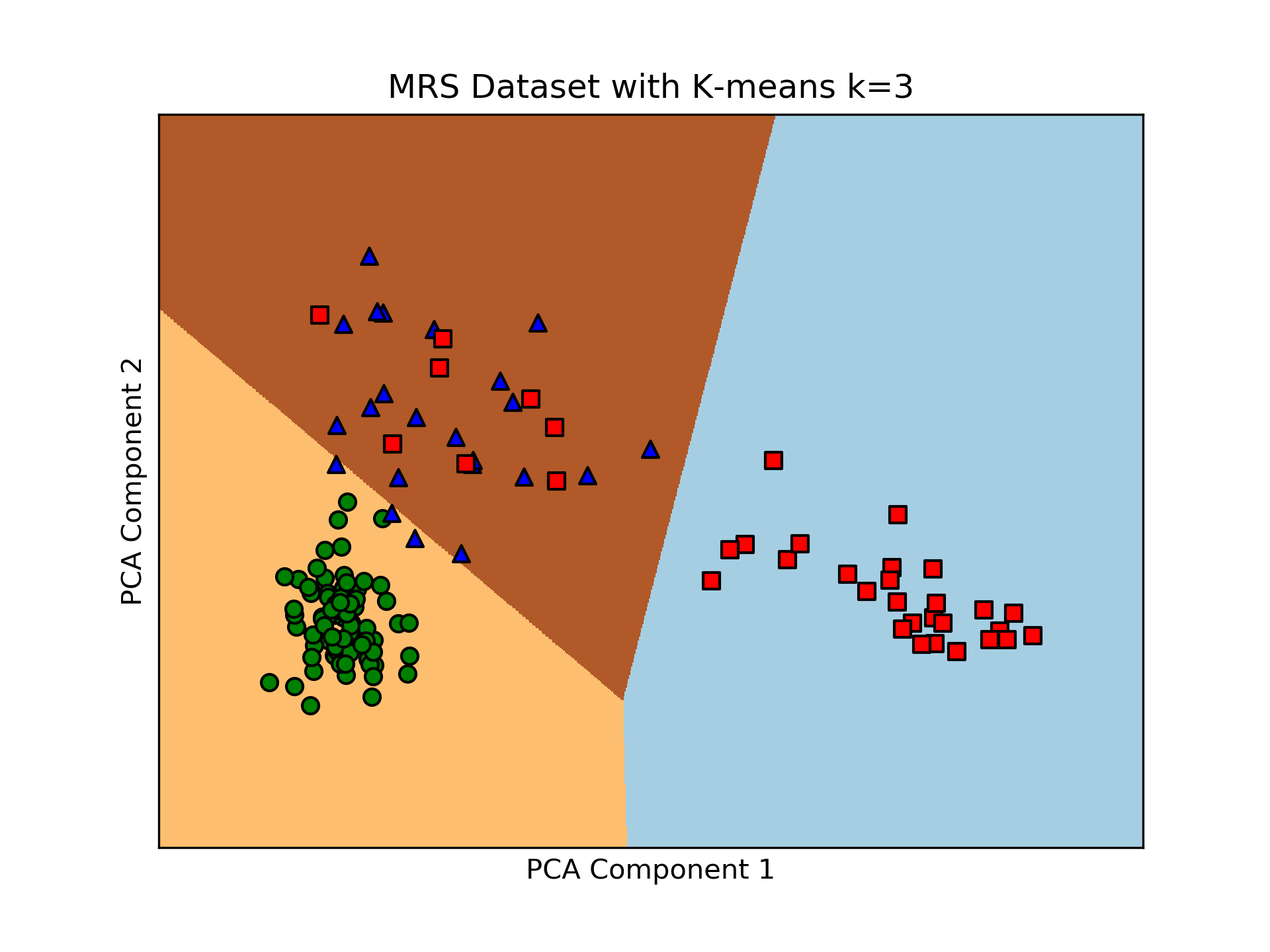}}
        \caption{Application of PCA to the MRS dataset with the first two components used~\ref{fig:datavis_pca} followed by k-means clustering~\ref{fig:datavis_kmeans}.}\label{fig:datavis}
    \end{figure}

    The nature of MRS data is high-dimensional (of roughly 1,024 dimensions) thus a reduction with Principal Component Analysis (PCA) was applied to explore the data, see Figure~\ref{fig:datavis}. Clustering with k-means results in a good separation of the classes, especially between healthy and high grade tissue (Figure~\ref{fig:datavis_kmeans}).
    
    \subsection{Methods}\label{sec:mandm_methods}
    
    This study took a three step approach. First, three different models (GAN, DCGAN and a generative adaptation of PMM) were used to generated synthetic MRS images based off of the training -MRS- dataset. Second, these samples were then each used to train a Random Forest to be able to classify MRS images as either healthy, low or high grade tissue. Finally, the results from the Random Forest classifiers were then analyzed and the synthetic data from the generative models were compared against the mean signal of the training dataset classes.
    
    The GAN model used in this study, based on~\cite{goodfellow_generative_2014}, had a generator comprised of a fully connected input layer which accepted a random value vector of size 100, with a hidden layer of 1,024 nodes in size using rectified exponential linear (ReLU)~\cite{nair_rectified_2010} as the activation function with the output layer having 1,024 nodes but utilizing TanH~\cite{agostinelli_learning_2014} as the activation function. The discriminator consisted of a linear input layer which took in an MRS image of 1,024 values, followed by a hidden layer consisting of 1,024 nodes using LeakyReLU~\cite{agostinelli_learning_2014} as the activation function, the output layer was a singular sigmoidal node~\cite{rumelhart_learning_1986} activation. The full model was trained on each class of tissue for over 150,000 epochs using the Adam optimizer~\cite{kingma_adam:_2014}.
    
    Compared to GANs, DCGANs use a deeper learning architecture which normally requires more data. To try and generate more expressive and domain accurate data, alterations were required to the training of the network to accommodate the limited data. The training process was modified to not take random batch samples but to deliberately take the full set of the data available for training at each epoch. The samples were then normalized with batch normalization. The architecture for the generator used in the study was a fully connected input layer comprised of 2048 nodes followed by four transposed one-dimensional convolutional layers using LeakyReLU as the activation function for all but the output layer which used TanH. The discriminator was the inverse with the first four layers being one-dimensional convolutions (using LeakyReLU as the activation function) followed by a single Sigmodial output node. Similar to the GAN model the DCGAN model was trained for at least 250,000 epochs on each tissue class.
    
    GANs are a non-linear method for the generation of data which, theoretically, should lead to more expressive data over that of more linear methods but to do so requires larger amounts of training data. To test the benefits, a comparison to a linear based generator in the study was required. For medical imaging data, factors such as patient orientation and relationship between values matters. Therefore, testing requires a method which can generate data that adheres to the domain. The selected method for this study was the Pairwise Mixture Model.
    
    The Pairwise Mixture Model (PMM)~\cite{olliverre_pairwise_2017} is a model for representing different brain tissue from MRS images and based on the work by \citeauthor{asad_supervised_2016}~\cite{asad_supervised_2016}, the purpose of which was to solve the problem of the heterogeneity of tissue types found within multi-voxel MRS images. The possible types were defined as normal, low (GI and GII type brain tumour tissue) and high (GIII and GIV brain tumour tissue). Each tissue model is expressed as a mean signal and the variation around the mean, calculated from applying PCA to a labelled dataset of homogeneous MRS images each relating to a specific grade. The models could then be as defined:
    
    \begin{equation}
        m_i(t) = \mu_i(t) + \sum \alpha_i e_{ik}(t),
    \end{equation}
    where \(\mu_i\) is the mean signal, \(\alpha_i\) and \(e_i\) respectively are the alpha weight coefficients and eigenvectors - which encode the variation around the mean signal - with K representing the number of eigenvectors determined for model \(m_{i}\). To calculate the amount of each tissue type found within a certain voxel the assumption that each voxel was a weighted sum of the possible tissue types gives the following:
    \begin{equation}
        s(t) = w_{n}m_{n}(t) + w_{l}m_{l}(t) + w_{h}m_{h}(t),
    \end{equation}
    which can then be viewed as an optimization problem where:
    \begin{align}
        E = \int [x(t) - s(t)]^{2} dt + \int [\sum_{j} s(t) - s_{j}(t)]^{2} dt, \\
        \text{with \(j\) representing the available surrounding model signals of \(s(t)\).} \nonumber
    \end{align}
    
    The estimated coefficients can be considered to represent the amount of each tissue type (normal, low or high) found within a voxel.
    
    By taking the models of the various tissue types from the PMM, it is straightforward to see how by varying the value of the coefficients of the models it is possible to create data that holds to the original domain. There is a limit to the amount of possibly created data (\( c^3 \) where \( c \) represents the coefficients) but it is still enough to train a classifier model with a deep learning architecture, based on the generated data alone.
    
    The testing of the generated data from each network used a set of simple, shallow Random Forest classifiers which were constructed and trained on the generated data from the GAN, DCGAN and the PMM generation method. A Random Forest was also trained on the training dataset alone as a control set. The Random Forest classifiers were then tested on the MRS image test set with the classification accuracy results recorded and examined. Furthermore the generated MRS images were compared to the mean signal from the training dataset for comparison to determine the adherence to the domain and possibility of expressiveness.

    
    \section{Results} \label{sec:results}
    
    Each model, the GAN, DCGAN and PMM generator, was trained on the MRS single-voxel training dataset for each class (normal, low and high). The GAN and DCGAN each had the architecture detailed in Section~\ref{sec:mandm_methods}.
    
    \begin{table}
    	\centering
        \caption{Table showing the resultant classification accuracy from the set of Random Forests trained on the generated data from the GAN, PMM and the DCGAN compared to that of a Random Forest trained on only the real, labelled training dataset (GT).}
        \begin{tabular}{lcccc}
            \noalign{\smallskip}
        	\hline\noalign{\smallskip}
        	Grade & PMM & GAN & DCGAN & GT \\
        	\noalign{\smallskip}
        	\hline
        	\noalign{\smallskip}
        	Healthy & 93\% & 97.5\% & 71\% & 96\% \\
        	Low & 96\% & 97\% & 66\% & 95\% \\
        	High & 93\% & 93\% & 71\% & 95\% \\
        	\hline
    	\end{tabular}
        \label{tab:results}
    \end{table}
    
    From the results shown in Table~\ref{tab:results} the GAN produced higher accuracy in the Random Forest classifier trained on its data than with the DCGAN generated method for low grade tumor patients and for normal (healthy) tissue. The classifier trained on PMM data produced higher accuracy for low grade tumor patients over that of the ground truth. The DCGAN produced the worst results over all tissue types.
    
    \begin{table}
    	\centering
        \caption{Table showing the mean squared error of the linear differences between the generated signals and the mean signal of the ground truth data for each class}
        \begin{tabular}{lccc}
            \noalign{\smallskip}
            \hline\noalign{\smallskip}
            Grade & PMM & GAN & DCGAN \\
            \noalign{\smallskip}
            \hline
            \noalign{\smallskip}
            Healthy & 0.002 & 1.021 & 5.148 \\
            Low & 0.004 & 2.044 & 4.063 \\
            High & 0.071 & 3.029 & 12.001 \\
            \hline
        \end{tabular}
        \label{tab:deltas}
    \end{table}
    
\begin{figure}
    	\centering
        \subcaptionbox{GAN Healthy. \label{gan_healthy}}{\includegraphics[width=0.25\textwidth]{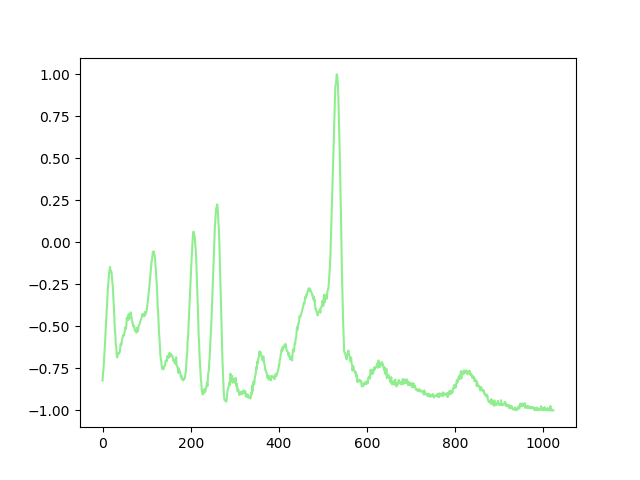}}
    	\subcaptionbox{GAN Low Grade. \label{gan_low}}{\includegraphics[width=0.25\textwidth]{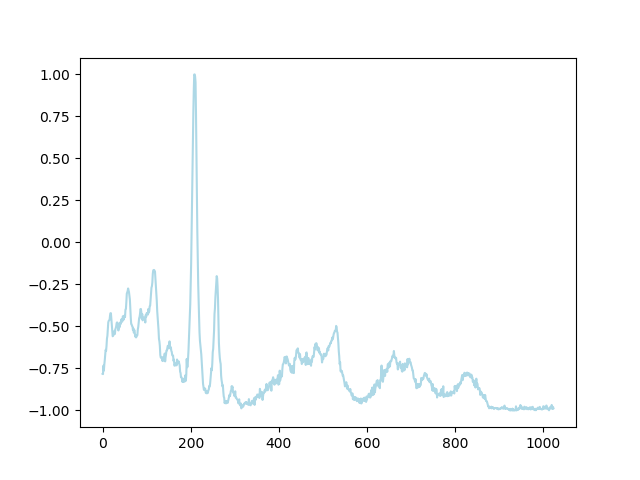}}
        \subcaptionbox{GAN High Grade. \label{gan_high}}{\includegraphics[width=0.25\textwidth]{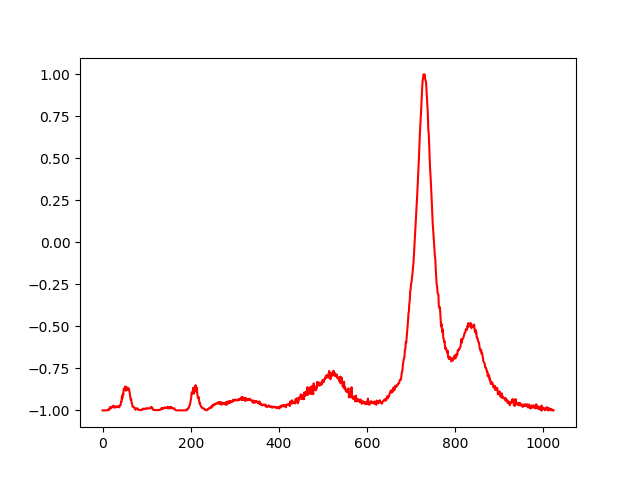}}
        \\
        \subcaptionbox{DCGAN Healthy. \label{dcgan_healthy}}{\includegraphics[width=0.25\textwidth]{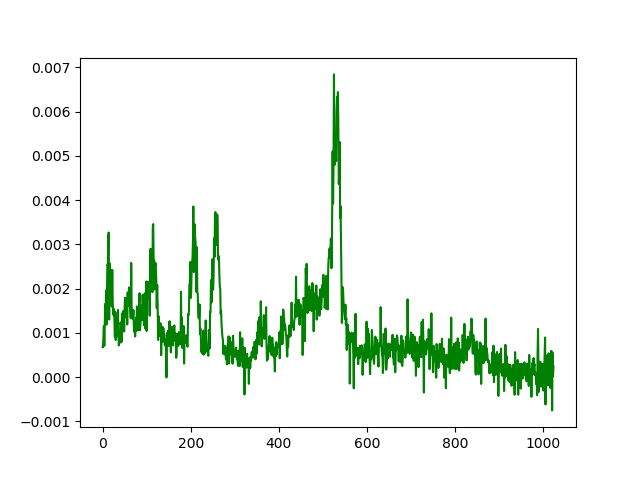}}
        \subcaptionbox{DCGAN Low Grade. \label{dcgan_low}}{\includegraphics[width=0.25\textwidth]{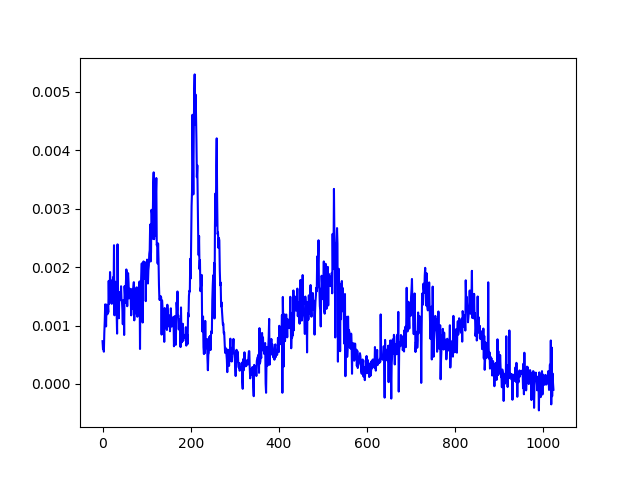}}
        \subcaptionbox{DCGAN High Grade. \label{dcgan_high}}{\includegraphics[width=0.25\textwidth]{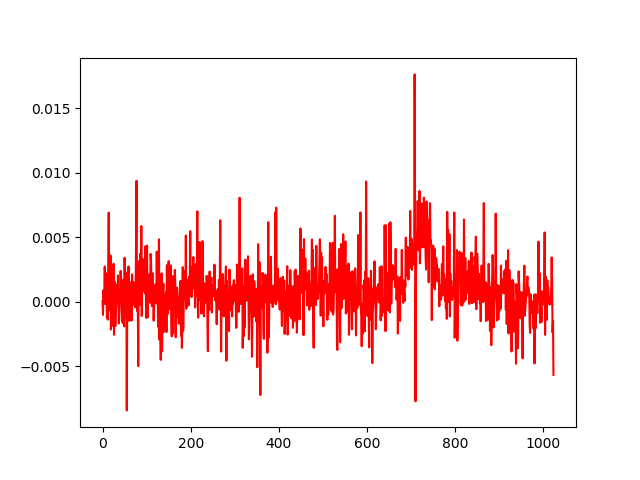}}
        \\
        \subcaptionbox{PMM Healthy. \label{pmm_healthy}}{\includegraphics[width=0.25\textwidth]{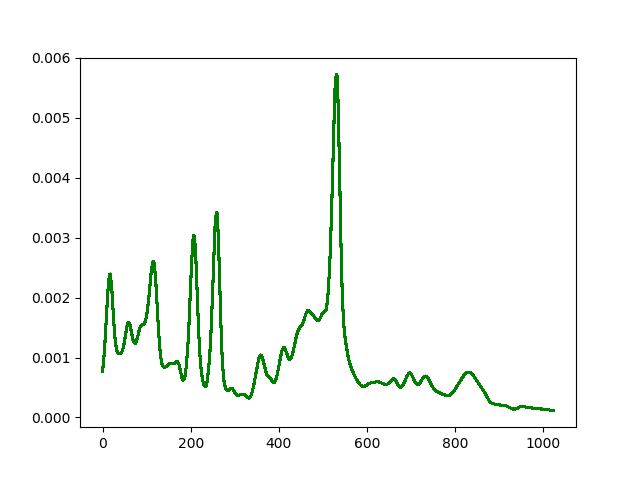}}
        \subcaptionbox{PMM Low Grade. \label{pmm_low}}{\includegraphics[width=0.25\textwidth]{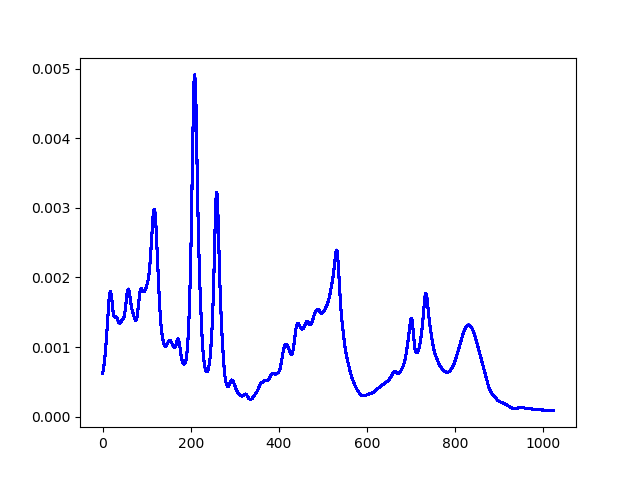}}
        \subcaptionbox{PMM High Grade. \label{pmm_high}}{\includegraphics[width=0.25\textwidth]{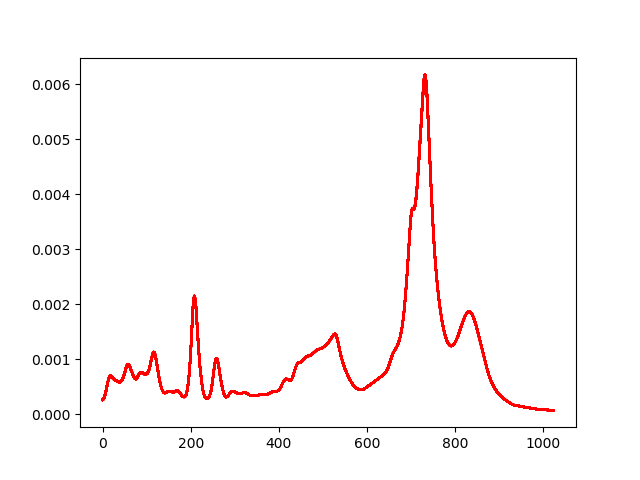}}
        \\
        \subcaptionbox{GT Healthy. \label{gt_healthy}}{\includegraphics[width=0.25\textwidth]{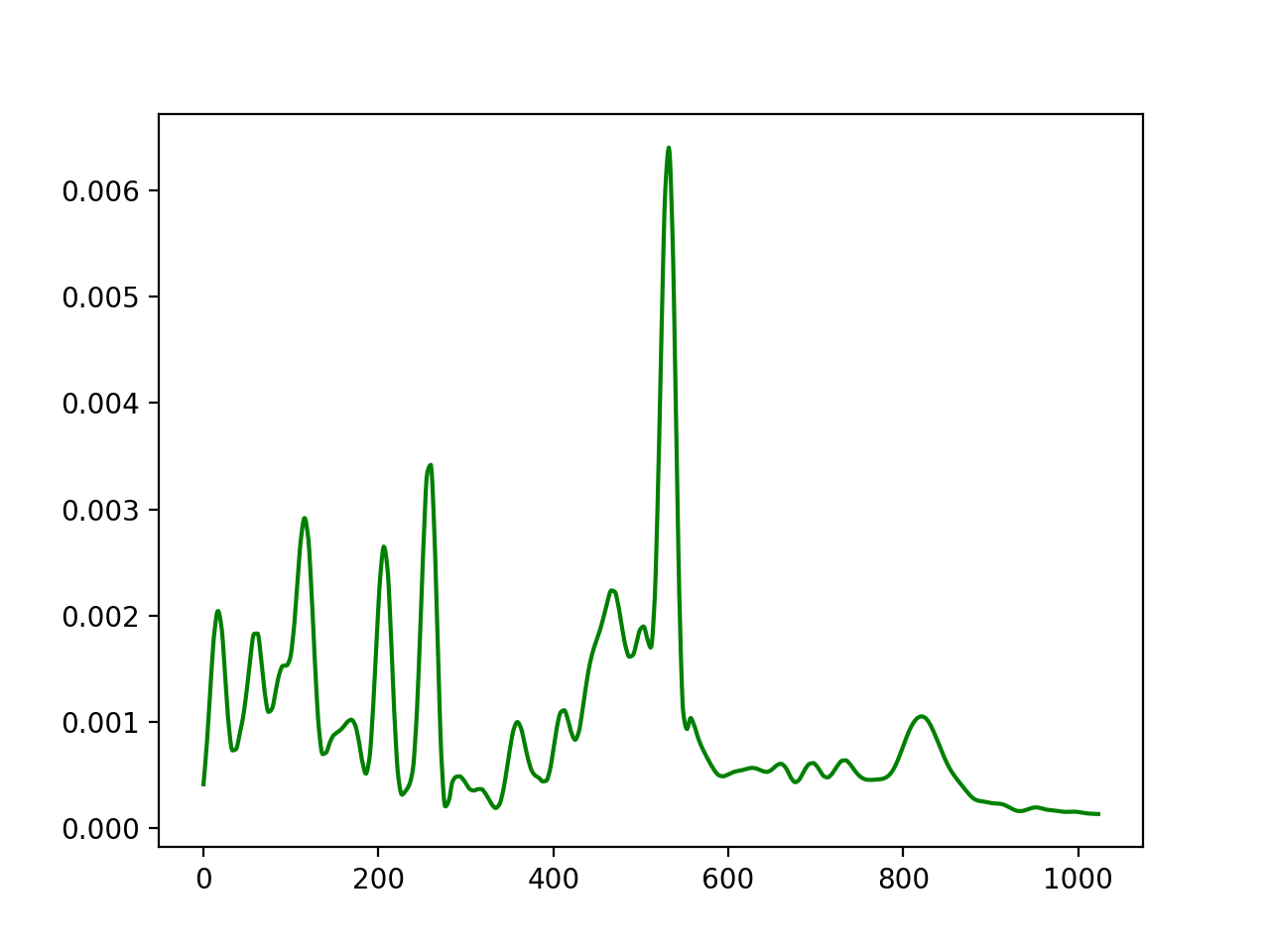}}
        \subcaptionbox{GT Low Grade. \label{gt_low}}{\includegraphics[width=0.25\textwidth]{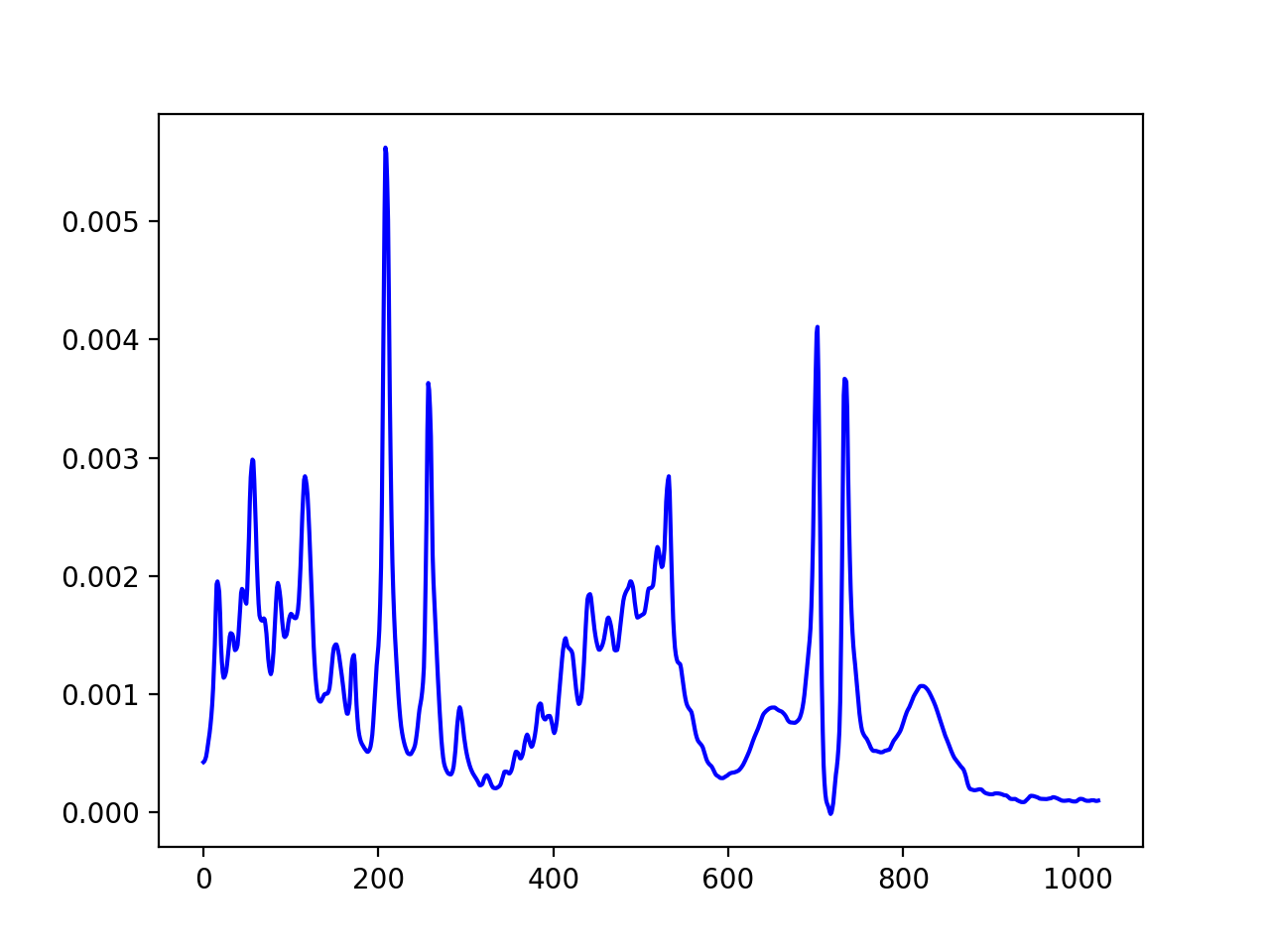}}
        \subcaptionbox{GT High Grade. \label{gt_high}}{\includegraphics[width=0.25\textwidth]{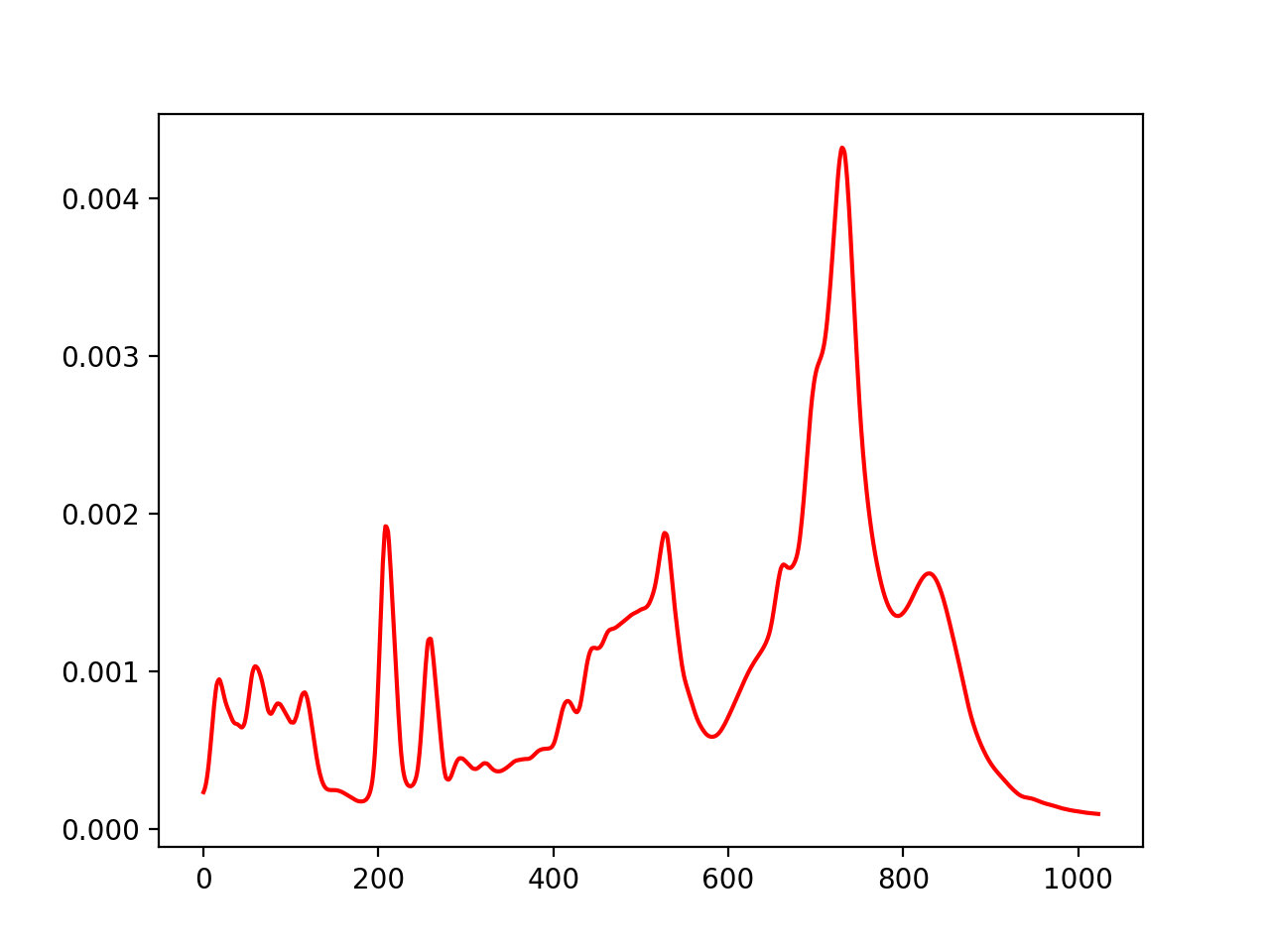}}   
        \caption{Artificially generated MRS signals of healthy (green), low grade (blue) and high grade (red) tumor tissue produced from the PMM generation method~\ref{pmm_healthy}~\ref{pmm_low}~\ref{pmm_high}, the GAN~\ref{gan_healthy}~\ref{gan_low}~\ref{gan_high} and the DCGAN~\ref{dcgan_healthy}~\ref{dcgan_low}~\ref{dcgan_high} compared with samples from the ground truth dataset~\ref{gt_healthy}~\ref{gt_low}~\ref{gt_high}.}
        \label{fig:mrs_outputs}
    \end{figure}

    The most variation within the resultant signals can be seen from both GAN methods, however, the GAN appears to cohere closer to the shape of the real MRS signals as shown in Figure~\ref{fig:mrs_outputs}, this can be shown through the deltas between signals shown in Table~\ref{tab:deltas}.
    
    \section{Conclusions} \label{sec:concl}
    
    This paper examines the possibility of generating MRS brain tumor images from limited and uneven data through one state-of-the-art generation technique, a deeper version of the same model and a modified MRS brain tumor model (GAN, DCGAN and PMM respectively). The results showed that the generated data could train a shallow Random Forest classifier to accurately determine the grades of brain tissue to the same level of one trained on real MRS data. The GAN trained model produced higher accuracy in testing and shows the expressiveness and capability in adversarial networks whereas the DCGAN model produced the lowest accuracy in classification as well as in similarity to the training data, highlighting the need for larger amounts of data when working with deep learning models. The linear model was able to produce spectra that were closer in appearance to the mean signal of the training dataset. The next stage is to examine the generated voxels by a domain expert to acknowledge their potential accuracy and expressiveness, the analysis drawn can then be used to determine what needs to be done in order to have the generated data successfully train a deep learning model for the classification of MRS images.

    
    \section*{Acknowledgements} \label{sec:acknw}
    
    We would like to thank Professor Franklyn Howe at St George's, University of London, for the brain tumour MR data used in this
    research as well as his insights into MRS. We are also grateful for the hardware provided to us by NVIDIA that was used in this research.

    \printbibliography
\end{document}